\documentclass[letterpaper, 10 pt, conference]{ieeeconf}  

\IEEEoverridecommandlockouts                              

\usepackage{graphicx} 
\usepackage{amsmath} 
\usepackage{placeins}
\usepackage{stfloats}
\usepackage{url}
\usepackage{booktabs}
\usepackage{titlesec}

\titlespacing\section{1pt}{6pt plus 4pt minus 2pt}{6pt plus 2pt minus 2pt}
\titlespacing\subsection{1pt}{6pt plus 4pt minus 2pt}{6pt plus 2pt minus 2pt}
\titlespacing\subsubsection{1pt}{2pt plus 4pt minus 2pt}{2pt plus 2pt minus 2pt}
\titlespacing\paragraph{1pt}{4pt plus 2pt minus 2pt}{4pt plus 2pt minus 2pt}

\title{\LARGE \bf
Learning and Control in Physically Coupled Dynamical Systems
}

\author{Bao Trung Truong \quad Dat Thanh Nguyen\quad Thinh T. Doan
\thanks{This work was supported in part by NSF under CAREER Award Grant No. 2527059 and Grant No. 2527044. The work of Bao Trung Truong and Dat Thanh Nguyen is also supported by the NSF CISE REU program under Grant No. CCF-2621730.}
\thanks{The authors are with the Department of Aerospace Engineering, The University of Texas at Austin, Austin, TX 78712 USA (e-mail: tbt589@my.utexas.edu;
tn22833@my.utexas.edu;
thinhdoan@utexas.edu)}%
}

\begin{document}

\maketitle
\thispagestyle{empty}
\pagestyle{empty}

\begin{abstract}
We introduce TAG (Testbed for Autonomous Games), a compact, low-cost robotic platform for studying learning and control in physically coupled dynamical systems. TAG combines an actively tilted and steered ball-in-maze board with a top-mounted rotary inverted pendulum. The pendulum's reaction torque and vibration disturb the board and marble, while the board's tilt, driven by the maze-solving policy, simultaneously disturbs the pendulum's balance. The main challenge in learning control policies for each subsystem is accounting for the impact of one subsystem's dynamics on the other's learning process. To address this challenge, we present complementary reinforcement learning paradigms, leveraging model-based and model-free approaches to learn the control policies for the maze and pendulum, respectively.  Our key finding is that introducing the pendulum causes an immediate, sharp drop in the maze policy's success rate, but the policy adapts online and recovers to its pre-coupling performance within a small number of learning episodes. Experimental results show that both policies, trained independently, successfully control the maze board and the pendulum despite their physical coupling.
\end{abstract}


\section{Introduction}
Many real-world robotic systems are not single, monolithic plants, but composed of physically coupled subsystems. These subsystems must be sensed, controlled, and often learned jointly. A prominent example is the humanoid robot, where legs and arms are physically coupled.  Locomotion must account for arm-manipulation dynamics, and vice versa. Designing control and learning algorithms for such coupled systems is difficult and costly. A poorly designed controller can damage the hardware, while exploratory actions during real-time learning can break robot components. This risk grows when a fault in one subsystem propagates through the coupling and destabilizes another. High-fidelity simulation offers a safer alternative, but policies trained purely in simulation often lose performance when transferred to real hardware  \cite{salvato2021crossing}. This sim-to-real gap remains a central challenge in robot learning. Thus, there is a need to develop a simple, low-cost, and safe testbed to study learning and control for physically coupled systems. Such a testbed would enable researchers to iterate and validate learning and control algorithms quickly on real hardware, before moving to more complex and expensive coupled robotic systems.

\begin{figure}[t]
    \centering
    \includegraphics[width=\linewidth]{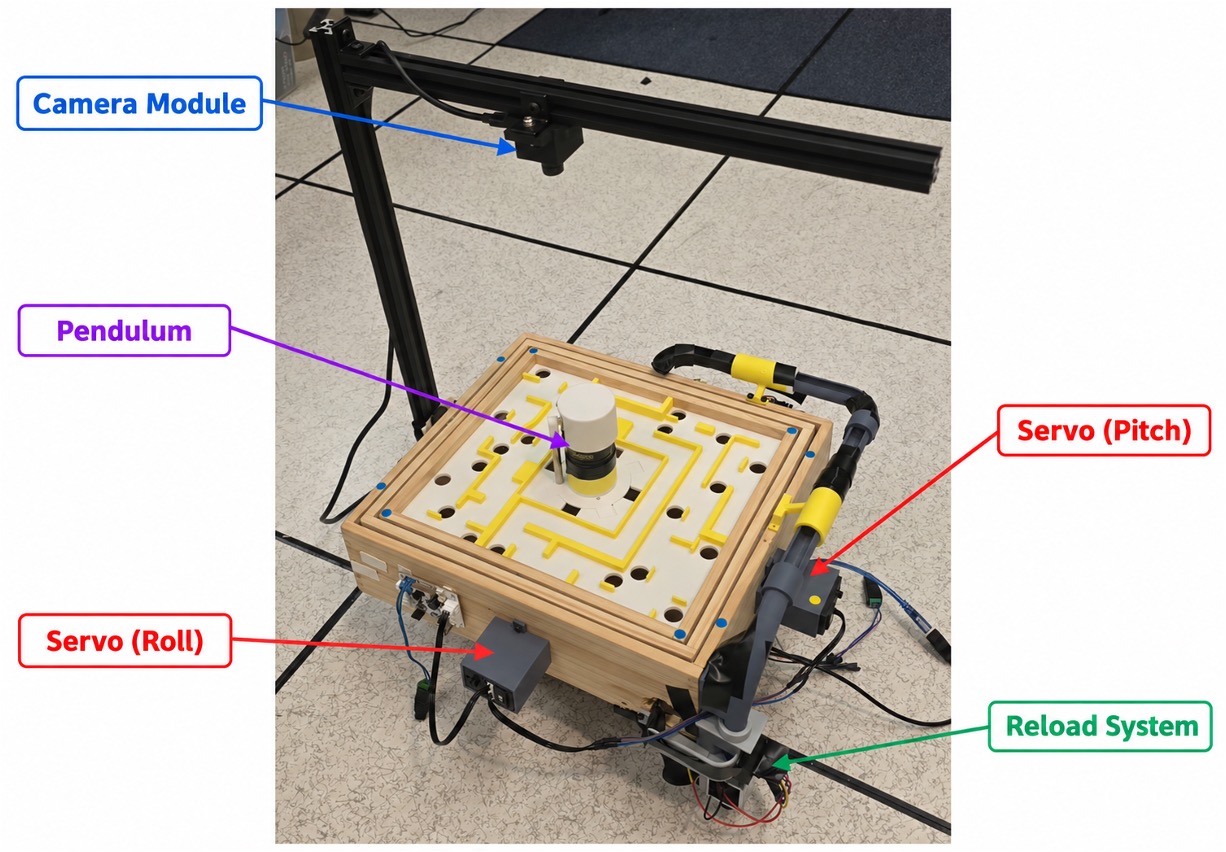}\vspace{-0.2cm}
\caption{\textbf{TAG:} A \textbf{T}estbed for \textbf{A}utonomous \textbf{G}ames}\vspace{-0.5cm}
    \label{fig:tag_hardware_overview}
\end{figure}


Motivated by this observation, we introduce a new compact robotic platform to study learning and control in physically coupled dynamical systems. We name our testbed \textbf{TAG}: a \textbf{T}estbed for \textbf{A}utonomous \textbf{G}ames, illustrated in Figure \ref{fig:tag_hardware_overview}. TAG combines two classic benchmark systems, a ball-in-maze board and a rotary inverted pendulum, into a single physically integrated system. The pendulum is mounted directly atop an actively tilted and steered maze board. As the pendulum balances, its reaction torque and vibration are transmitted directly into the board's dynamics, and in turn into the motion of a marble navigating the maze under overhead visual tracking. Conversely, the board's own movement, driven by the maze-navigation task, perturbs the pendulum's balance. Each subsystem has its own controller, and its design must account for the presence of the other. 


To learn control policies for TAG, we develop complementary model-based and model-free learning algorithms, one for each subsystem. The maze-navigation subsystem uses DreamerV3 \cite{hafner2023dreamerv3}, a model-based algorithm well suited to costly real rollouts. DreamerV3 was previously used for autonomous maze-solving in CyberRunner \cite{bi2024sample}, but without coupling to a second physical subsystem. We build on this by extending to a physically coupled system. For the pendulum, we instead use the model-free TQC algorithm \cite{kuznetsov2020tqc}, trained entirely in simulation and deployed as a fixed policy on hardware. Our experiments test whether these two separately optimized controllers remain effective when operating together on the same mechanically coupled platform. Our experimental results show that the pendulum policy sustains high success across a wide range of board disturbances, degrading gracefully and being more sensitive to disturbance dimensionality than to speed. On the other hand, the maze policy suffers a sharp but transient drop upon pendulum activation, recovering within a small fraction of the episodes required for initial learning. Our key finding is that independently-trained policies, with no shared model of one another, can each remain effective despite continuous, physically transmitted interference from the other.\vspace{-0.1cm}



\subsection{Related Works}\vspace{-0.1cm}
\noindent\textbf{Physically coupled robotic systems.} A growing body of work studies robots whose subsystems are mechanically coupled and must be controlled jointly. Quadrotors carrying suspended payloads exhibit strongly coupled, underactuated dynamics that motivate controllers modeling the full vehicle-payload system \cite{sreenath2013geometric}. In legged manipulation, pairing a model-based arm controller with a separate RL-based locomotion controller, treating manipulator motion purely as a disturbance to the base, lets each subsystem perform well individually, but the two controllers can only passively adapt to one another \cite{ma2022combining}. Humanoid robots exhibit a more bidirectional coupling: momentum from leg motion affects whole-body dynamics, while coordinated arm motion can generate opposing momentum, improving disturbance recovery over a fixed-arm baseline \cite{lee2025learning}. Free-floating space manipulators show a still more fundamental case, where momentum conservation dictates that manipulator motion necessarily induces spacecraft attitude reaction, a coupling studied since \cite{dubowsky1991path} and sometimes exploited deliberately for propellant-free attitude control rather than merely rejected. TAG is motivated by the same phenomenon, but isolates it in a compact, fully-observable tabletop platform rather than a full-scale legged, aerial, or orbital system, making the coupling itself easier to vary and study systematically.

\noindent\textbf{Benchmark testbeds for learning-based control.} Ball-in-maze and rotary inverted pendulum systems are long-standing benchmarks for control and reinforcement learning individually. Maze-navigation benchmarks typically use a force-actuated ball reaching a target within a closed maze \cite{fu2020d4rl}, while ball-in-maze systems on tip-tilt platforms have separately been proposed as physical RL benchmarks for their rich dynamics and ease of construction \cite{romeres2019semiparametrical}. The rotary inverted pendulum is a standard underactuated benchmark, used to evaluate both model-free and model-based RL for swing-up and stabilization \cite{amadio2022model,hong2023optimizing}. In both lines of work, disturbances are typically exogenous and unstructured, rather than the controllable byproduct of a second, independently actuated subsystem. More broadly, most learning-based control is validated primarily in simulation, and resulting policies often fail to transfer due to the sim-to-real gap \cite{salvato2021crossing}, motivating low-cost, reproducible physical benchmarks such as robotic arm cells \cite{yang2019replab}, legged platforms \cite{ahn2020robel}, and miniature driving benchmarks \cite{o2020f1tenth}. TAG follows this same low-cost, reproducible philosophy, while specifically targeting physically coupled dynamics rather than a single subsystem.

\noindent\textbf{Model-based and model-free learning for control.} We will leverage complementary reinforcement learning paradigms. The model-based Dreamer family learns a compact latent world model and trains policies through imagined rollouts \cite{hafner2023dreamerv3}. DreamerV3 has been applied to autonomous maze-solving on a physical ball-in-maze platform \cite{bi2024sample}, which we build on and extend to a physically coupled setting. Model-free off-policy actor-critic methods are widely used for underactuated balancing \cite{hong2023optimizing}; we specifically use Truncated Quantile Critics (TQC), whose distributional, truncated critic is designed to reduce overestimation bias in continuous control \cite{kuznetsov2020tqc}. Together, these paradigms let us match each subsystem's learning algorithm to the structure of its dynamics, without either policy requiring a model of the other.

\section{TAG: Hardware and Software Designs}
\label{sec:hardware}


\subsection{Ball-in-Maze Boards} \vspace{-0.1cm}
We adopt the ball-in-maze hardware design from the CyberRunner project
\cite{bi2024sample}, with modifications to match our system. The playing
board is a custom $259 \times 229$\,mm surface with a fixed route bounded
by walls and interrupted by holes; roll and pitch servos tilt the board
about two perpendicular axes, letting the policy steer the marble along
the route, while an automatic reload system returns the marble to its
starting position after each episode. An episode succeeds if the marble
reaches the goal and terminates early if it enters a hole. Further
hardware details are given in \cite{bi2024sample}.

State estimation follows the same overhead-camera pipeline: a See3CAM
camera captures the playing surface at $1280\times720$, downsampled and
padded to $640\times400$, from which a learned ONNX detector locates the
marble and a perspective-n-point calculation, using markers on both the
moving board and the stationary outer frame, recovers board pose relative
to a fixed world reference. This yields marble position $(x_b, y_b)$, velocity $(\dot{x}_b, \dot{y}_b)$, and board angles $(\alpha,
\beta)$ in a common board frame, used directly by the maze controller.\vspace{0.1cm}

\noindent\textbf{Computing Resources.} The maze system splits computation
between a local computer and a remote GPU server. The local computer, a
laptop with an Intel Core i7 and an NVIDIA RTX 4070 Laptop GPU, handles
all hardware-side processing, including camera acquisition, marble
detection, board-pose estimation, and servo communication, coordinated
locally via ROS~2. Observations are sent over TCP to a remote NVIDIA RTX
6000 Ada server, which runs DreamerV3 for learning and inference and
returns roll/pitch actions to the local computer for execution.

Because observations and actions cross this network link, the control
loop includes both computation and communication delay. These delays are
present throughout real-system training, so the policy learns under the
same observation-action latency it experiences at evaluation. Running the
computationally intensive DreamerV3 workload remotely lets sensing,
actuation, and safety functions remain local and responsive; moving
learning and inference to a sufficiently powerful local workstation would
eliminate most network delay and is a natural direction for a faster
control loop.\vspace{-0.1cm}

\subsection{Inverted Pendulum}\vspace{-0.1cm}
\begin{figure}[h]
    \centering
    \includegraphics[width=0.4\textwidth]{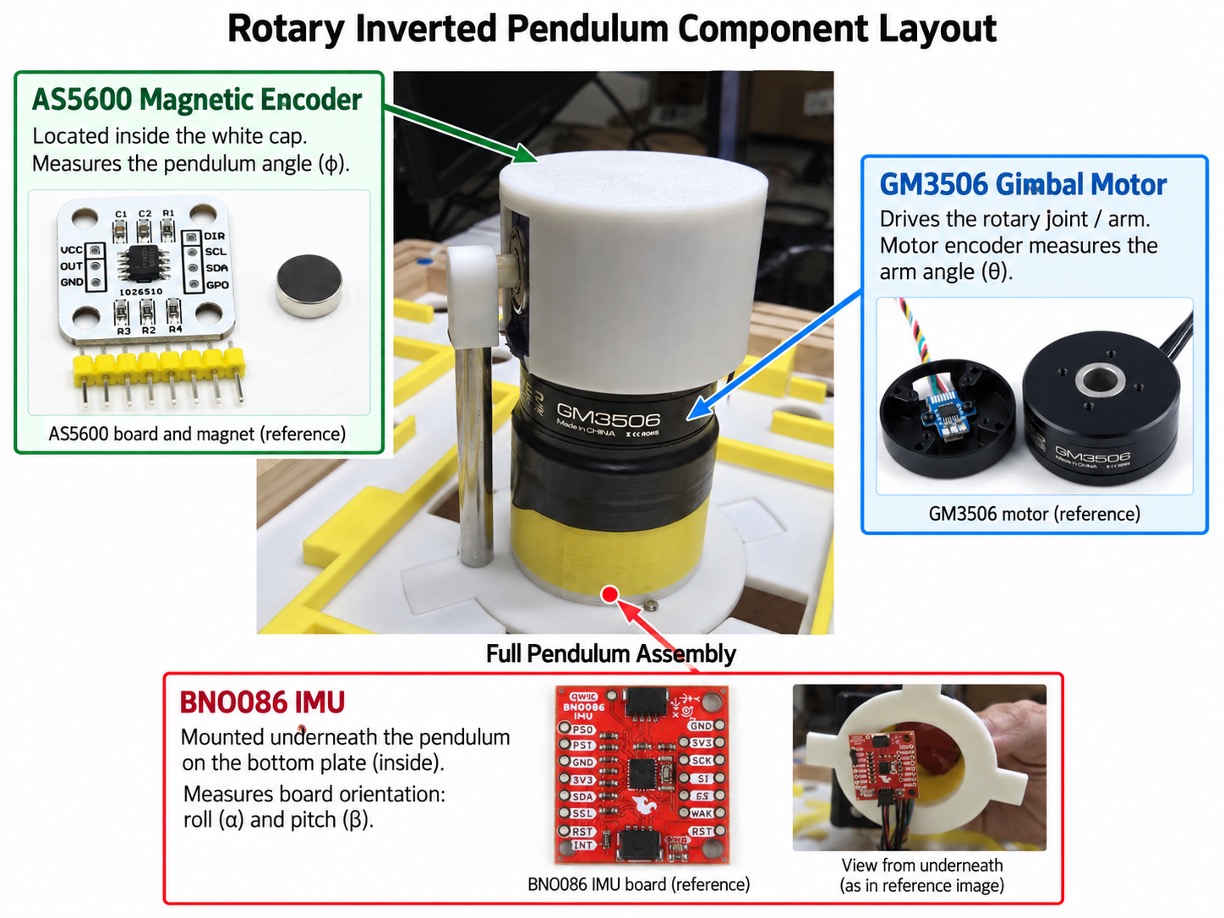}\vspace{-.2cm}
    \caption{Rotary inverted pendulum.}
    \label{fig:pendulum_assembly}\vspace{-0.2cm}
\end{figure}
A rotary inverted pendulum (Figure \ref{fig:pendulum_assembly}) is mounted at the center of the tilting board. The rotary arm is driven by a BLDC/gimbal motor through a TMC6300 driver under open-loop field-oriented control, with a passive joint swinging freely at the arm's end. Since the pendulum base is rigidly attached to the board, motor torque feeds back into the board as reaction torque, and the spinning mass adds higher-frequency vibration. This coupling is bidirectional: pendulum motion disturbs the marble, while the board's roll and pitch continuously affect the gravity direction of the pendulum.

Two independent sensing chains support the coupled system. The pendulum joint angle is measured by an AS5600 magnetic encoder (4096 counts/revolution), and the rotary-arm angle by a separate AS5048A magnetic encoder over SPI, together giving the pendulum's full joint state and, by differentiation, its angular velocities. Board orientation and rate are measured independently by a SparkFun BNO086 IMU mounted on the moving board, reporting roll, pitch, and gyro rates at roughly 100 Hz against a 200 Hz control loop. Pendulum sensing, state estimation, and policy inference all run onboard an embedded ESP32. Here, the policy's actors are exported from the same remote GPU server.

\noindent\textbf{MuJoCo Simulators for Pendulum.} 
All pendulum training is carried out in simulation. We model the plant in MuJoCo as a Furuta
pendulum rigidly mounted on a two-axis gimbal that reproduces the board's roll
and pitch, so the same base motion the maze induces is applied to the pendulum
in simulation. The rotary arm is driven by a torque actuator implementing the
identified voltage-to-torque model and delay chain of the following subsection,
while the two board axes are position-controlled joints. The task is wrapped as
a Gymnasium environment that integrates MuJoCo at a fixed
$1$~ms step, decimated to the $5$~ms ($200$~Hz) period at which the policy
observes and acts.\vspace{-0.4cm}
\begin{figure}[thpb]
    \centering
    \includegraphics[width=0.8\columnwidth]{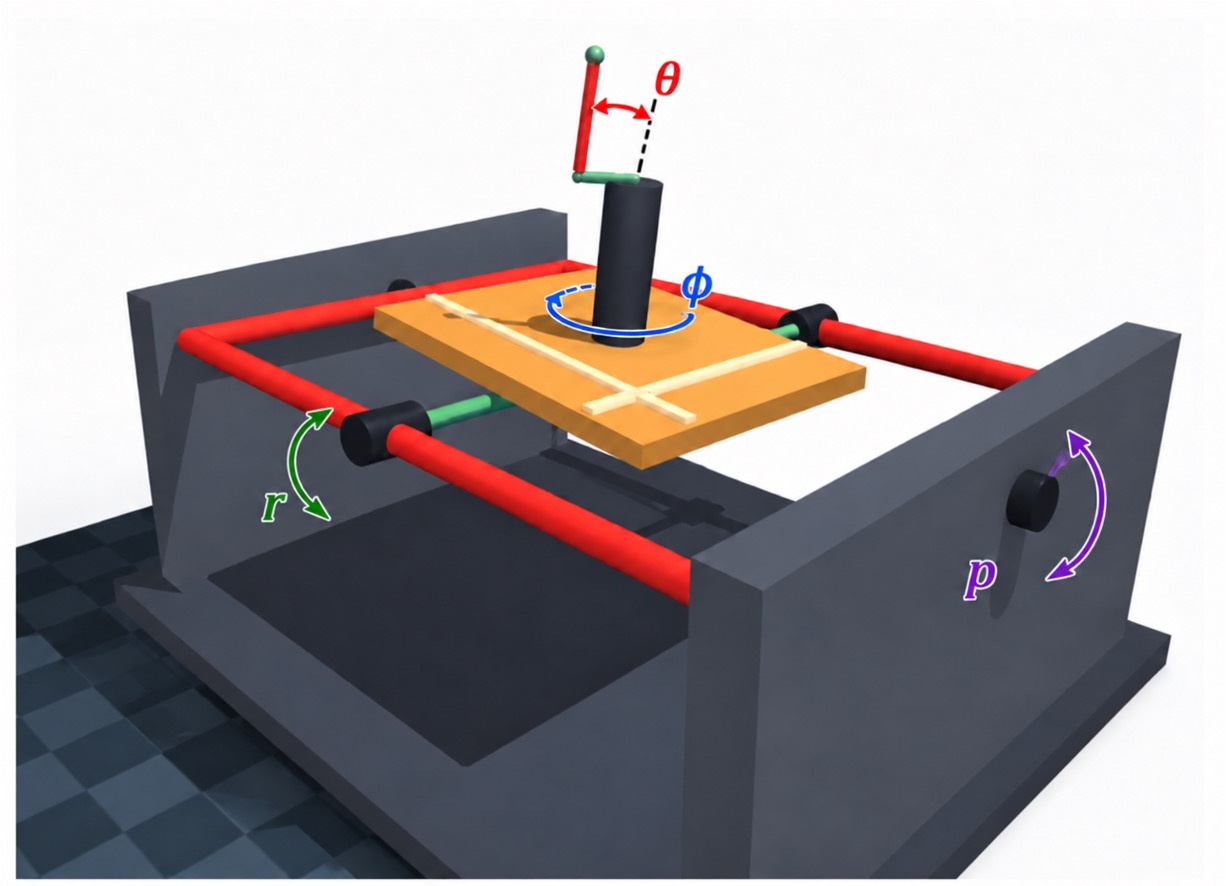}\vspace{-0.2cm}
    \caption{MuJuCo Simulator for Training Pendulum.}
    \label{fig:furuta_coords}\vspace{-0.4cm}
\end{figure}
\section{Learning and Control Methods}\vspace{-0.1cm}
\label{sec:methods}
TAG runs two learned policies simultaneously on the same physical structure: one commands the board's roll and pitch to steer the marble through the maze, and the other commands the rotary-arm motor voltage to keep the pendulum upright. Neither controller has access to the other's internal state, reward, or training process, so each policy must learn, independently, to remain effective despite disturbances transmitted through the shared board. This section presents our approach to learning both policies, leveraging the complementary advantages of model-based and model-free reinforcement learning (RL) to match each algorithm to the structure of its subsystem's dynamics.

\begin{figure*}[t]
    \centering
    \includegraphics[width=\textwidth]{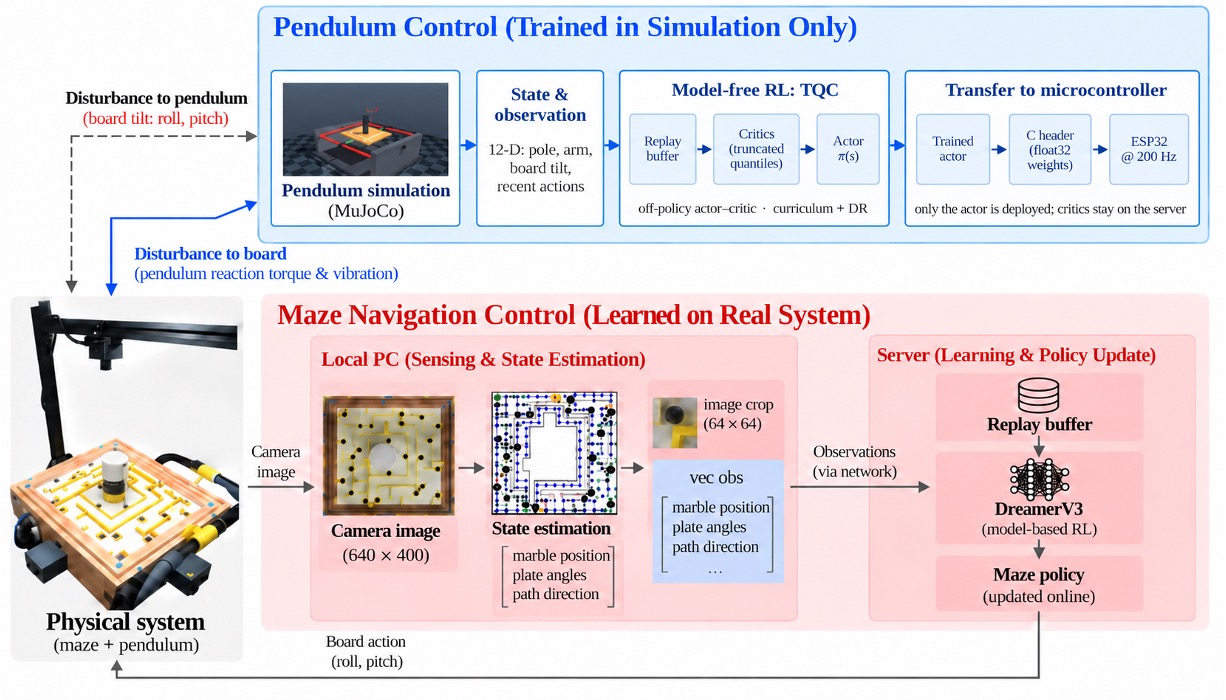}\vspace{-0.3cm}
    \caption{\textbf{Overview of the proposed method.} The pendulum policy is trained in simulation with a model-free algorithm and deployed fixed on hardware; the maze policy is trained online, directly on the physical system, using DreamerV3. Camera images are processed locally into observations, sent to a remote server that runs DreamerV3 and updates the maze policy via a replay buffer, and returned as roll/pitch commands to the board. The two subsystems are physically coupled through the shared board, with the pendulum acting as a disturbance on the maze policy.}\vspace{-0.5cm}
    \label{fig:training_flow}
\end{figure*}
As summarized in Fig.~\ref{fig:training_flow}, the maze-navigation policy is learned with a model-based algorithm, DreamerV3, which optimizes the policy through imagined rollouts in a compact latent space learned directly from observations. This policy is trained and continually adapted online, directly on the physical hardware, using data collected during real operation. The pendulum-stabilization policy, in contrast, is learned with a model-free algorithm, trained entirely in simulation and deployed to hardware without further updates. This asymmetry follows directly from how well each subsystem's dynamics can be simulated: the marble's contact-rich interaction with the maze surface, and the visual pipeline used to observe it, are difficult to model accurately enough for training in simulation alone, whereas the pendulum's rigid-body dynamics are well understood and can be identified and simulated accurately, making simulation-only training and zero-shot hardware deployment viable. This asymmetry, online adaptation from real-world experience versus offline training and zero-shot deployment, reflects the advantages of each algorithm and subsystem dynamics.\vspace{-0.1cm}

\subsection{Pendulum Control}\vspace{-0.1cm}

We formulate the pendulum-stabilization task as a continuous-state,
continuous-action MDP. Because several relevant quantities (pendulum
angle, board tilt, and their rates) are not directly usable in raw sensor
form, we define an observation vector $\mathbf{s}_t$, distinct from the
underlying physical state, that is transformed and normalized for
learning.

\noindent\textbf{State Space.} A state is composed of $12$ components:
\begin{equation*}
\begin{aligned}
\mathbf{s}_t = \Big[&\cos\theta_t,\ \sin\theta_t,\ \tfrac{\dot\theta_t}{15},\ \tfrac{\phi_t}{\pi},\ \tfrac{\dot\phi_t}{25},\ a_{t-1},\\
&\tfrac{r_t}{15^\circ},\ \tfrac{p_t}{15^\circ},\ \tfrac{\omega_{x,t}}{80^\circ/s},\ \tfrac{\omega_{y,t}}{80^\circ/s},\ a_{t-2},\ a_{t-3}\Big].
\end{aligned}
\label{eq:observation}
\end{equation*}
The pendulum angle $\theta$ is encoded
as $(\cos\theta, \sin\theta)$ rather than raw, avoiding the discontinuity where the pendulum spends most of its
time, near upright. Board roll $r_t$, pitch $p_t$, and their rates
$\omega_{x,t}, \omega_{y,t}$ are observed directly, since they carry the
disturbance injected by the maze policy, letting the controller act directly on
the disturbance. The three most recent actions $a_{t-1}, a_{t-2}, a_{t-3}$
compensate for actuator delay: since a command issued several steps
earlier may still be taking effect, observing recent action history lets
the policy track pending actuator response. Every term is
scaled to roughly its physical range (e.g., an arm rate near 25~rad/s
maps close to 1), identically on the embedded controller, so nothing
shifts between training and deployment.

\noindent\textbf{Action Space.} The action $a_t$ is
a single continuous voltage command to the rotary-arm motor, a
BLDC/gimbal motor driven through a TMC6300 driver under open-loop
field-oriented control. Commanded voltage produces torque as
\begin{equation*}
\tau_t = K_M V_{motor,t} - K_E\dot\phi_t - b_\phi\dot\phi_t - \tau_c\,\mathrm{sgn}(\dot\phi_t),
\label{eq:motor_torque}
\end{equation*}
where $V_{motor,t}$ is $a_t$ rescaled to the driver's voltage range,
$\dot\phi_t$ is arm angular velocity, and $K_M$, $K_E$, $b_\phi$,
$\tau_c$ are the identified motor-gain, back-EMF, viscous-damping, and
Coulomb-friction constants.

\noindent\textbf{Reward Design.} Rewarding only for passing through
upright is insufficient, since a policy can whip through vertical too
fast to settle there. We instead shape a stage cost as
\begin{equation*}
\begin{aligned}
J_t = \underbrace{(1-u_t)}_{\text{upright error}} + \underbrace{w_\phi\!\left(\tfrac{\phi_t}{\pi}\right)^{\!2}}_{\text{arm centering}} + \underbrace{0.005\,a_t^2}_{\text{effort}} + \underbrace{0.002\,\dot\phi_t^2}_{\text{arm speed}} \\
+ \underbrace{w_{\Delta a}(a_t-a_{t-1})^2}_{\text{smoothness}} + \underbrace{0.01\,\mathbf{1}(u_t{>}0.5)\dot\theta_t^2}_{\text{pole settling}} \\
+ \underbrace{P_{arm}(\phi_t)}_{\text{arm envelope}} + \underbrace{P_{cable}(\phi_t)}_{\text{cable safety}} - \underbrace{B_{upright}}_{\text{stability bonus}},
\end{aligned}
\label{eq:reward}
\end{equation*}
where $u_t = \hat z_{pole}^\top \hat z_{world} = \cos(e_t)$ measures
alignment with true gravity-vertical in the world frame ($u_t{=}1$
upright), so the target direction never shifts as the board tilts; a
board-frame definition would instead conflate the pendulum's objective
with the disturbance acting on it.

The pole-settling term activates only once $u_t$ exceeds 0.5, penalizing
$\dot\theta_t$ near upright but not during swing-up, when angular
velocity is needed to gain energy; this lets a single reward express both
build energy and then kill it without an explicit mode switch. Cable
safety is similarly staged: $P_{cable}(\phi_t)$ ramps in as a soft
penalty once the arm exceeds $270^\circ$ of rotation, pushing the policy
back toward center well before real risk, while a hard $330^\circ$ limit
terminates the episode with a fixed $-10$ penalty as a rare backstop
rather than a routine constraint.

\noindent\textbf{Model-Free RL Methods.} 
The pendulum must be swung up from rest and stabilized at true gravity-vertical using a single actuator, the rotary arm, while the board beneath it is continuously tilted by the maze policy. Because the pendulum joint itself is passive, energy must first be pumped into the system through the arm before the pendulum can be caught and balanced near its unstable equilibrium. Classical approaches address this by hand-splitting the task into a separate energy-based swing-up controller and a balancing stage (e.g., LQR or PID), switched based on pendulum  \cite{aastrom2000swinging}. We instead train a single policy with Truncated Quantile Critics (TQC) \cite{kuznetsov2020tqc}, a model-free off-policy actor-critic method, letting it learn the swing-up-to-balance transition directly from experience, without hand-coded switching logic.

The maze policy is not present during pendulum training, since the two subsystems are trained independently. The board disturbance is supplied by a randomized tilt generator emitting continuous roll/pitch trajectories spanning the same two-axis envelope used at deployment. Sensor noise, bias, and the $100$~Hz IMU reporting rate against the $200$~Hz control loop are simulated to match deployment conditions.

Training proceeds through a curriculum learning \cite{bengio2009curriculum} rather than directly on the full disturbance range, since a randomly initialized policy trained against full 
$\pm 15^{\circ}$ two-axis motion rarely reaches upright and provides little learning signal. We first train and verify a single-axis swing-up-and-balance policy (confirmed at approximately $91.5$\% success), then expand its input layers to accept a second axis and continue training through progressively harder stages, from a stationary board to full two-axis disturbance at increasing speed. At each stage transition, the actor is preserved, since it already encodes useful behavior, while its critics are reset and given a brief actor-frozen warm-up period, since they were fit to the previous, now-outdated disturbance level; a teacher-anchoring loss term further discourages the policy from drifting away from previously-learned behavior while the new critics stabilize. Each stage is screened over 300 episodes per condition before advancing, and the final policy is re-verified over 500 episodes per condition with reported confidence intervals; success requires holding the pendulum within its upright and arm-angle limits for at least 80\% of a training evaluation window.

To close the sim-to-real gap, the nominal simulation model is built from hardware-identified actuator parameters rather than assumed values. In addition, during later curriculum learning stages, we randomize motor gain, damping, friction, pendulum inertia, observation noise, and action delay around these identified values \cite{peng2018sim}, see Table \ref{tab:dr}, so the deployed policy has been trained against a family of plausible plants rather than a single fixed simulation. Only the trained actor, not the critics, is exported to a C header and run onboard the ESP32 at 
200
200~Hz using identical observation ordering and normalization as in training.

\begin{table}[h]
\caption{Domain randomization for pendulum training}
\label{tab:dr}
\centering
\footnotesize
\setlength{\tabcolsep}{4pt}
\begin{tabular}{@{}lcc@{}}
\toprule
Parameter (unit) & Nominal & Range \\
\midrule
Motor gain $K_M$ (N$\cdot$m/V)                    & $0.0127$ & $[0.010,\,0.016]$ \\
Arm damping $b_\phi$ ($10^{-4}$ N$\cdot$m$\cdot$s/rad) & $9.4$  & $[3.0,\,10.0]$ \\
Arm Coulomb friction (mN$\cdot$m)                 & $6.0$    & $[4.0,\,8.0]$ \\
Pole damping $b_\theta$ ($10^{-5}$ N$\cdot$m$\cdot$s/rad) & $3.4$ & $[1.4,\,5.4]$ \\
Pole Coulomb friction (mN$\cdot$m)                & $0.165$  & $[0.065,\,0.265]$ \\
Pole inertia (scale)                              & $1.00$   & $[0.92,\,1.08]$ \\
Joint-angle obs.\ noise, std (rad)                & $0$      & $[0,\,0.01]$ \\
Actuator lag (ms)                                 & ---      & $[3,\,9]$ \\
Action delay (steps, $5$--$10$\,ms)               & $1$      & $\{1,\,2\}$ \\
\bottomrule
\end{tabular}\vspace{-0.1cm}
\end{table}

\subsection{Maze Board Control}\vspace{-0.1cm}
The maze policy is trained with DreamerV3, which learns a recurrent world model from replayed experience and trains an actor and critic entirely on
imagined trajectories inside that model, rather than on real rollouts \cite{bi2024sample}.
This is due to real rollouts are costly: an
autoload mechanism resets the marble after every episode, but the servos
still heat up under continuous operation and must be periodically rested.

\noindent\textbf{System Architecture.} The maze hardware runs the camera driver, state estimator, and servo
driver as ROS~2 nodes in real time, while a separate GPU machine runs the
DreamerV3 learner, connected over a TCP bridge; the hardware streams
observations out and receives actions back at whatever rate the network
and learner allow. This decouples real-time sensing and actuation from
training, so the learner can fall behind, restart, or run elsewhere
without touching the robot's control loop, and a trained checkpoint can
later be swapped in for evaluation without re-plumbing.

\noindent\textbf{Observation.} At every control step,
\begin{equation*}
o_t = \{I_t,\ \alpha_t,\ \beta_t,\ x_t,\ y_t,\ v_{x,t},\ v_{y,t},\ g_t,\ p_t,\ d_t\},
\end{equation*}
where $I_t$ is a locally cropped image of the maze around the marble,
$\alpha_t,\beta_t$ are board pitch and roll, $(x_t,y_t)$ and
$(v_{x,t},v_{y,t})$ are marble position and velocity, $g_t$ is a short
window of upcoming reference-path points, $p_t$ is scalar path progress,
and $d_t$ is an optional pendulum-state feature block populated only once
pendulum-active training begins. Velocity is included explicitly because
position alone cannot distinguish a marble decelerating into a hole from
one about to roll past it. Marble position comes from an ONNX detector
rather than color thresholding, since the marble's appearance shifts too
much under the tilting board's glare and shadow; color tracking is
instead used for eight reference markers, four fixed to the outer frame
and four on the tilting plate, giving a fresh pixel-to-metres homography
each frame and a stable world reference for board angle.

The policy outputs $a_t \in [-1,1]^2$, rescaled to pitch and roll
setpoints sent to the servos as position commands; a rate limiter may
smooth abrupt changes, but the target angle itself is chosen entirely by
the policy, leaving low-level actuation tracking to the servo's own
position loop.

\noindent\textbf{World Model.} A CNN encodes $I_t$, an MLP encodes the
numerical part of $o_t$, and both feed a recurrent state-space model
(RSSM) carrying hidden state across control steps, capturing information
a single frame cannot, such as true marble velocity, servo lag, or the
current phase of pendulum disturbance. Small MLP heads predict the next
latent state, reward, and episode continuation from this hidden state,
and the actor and critic are trained entirely on rollouts imagined from
these predictions.

\noindent\textbf{Reward.} Reward is primarily a progress signal,
\begin{equation*}
r_t = \begin{cases}
0, & \text{denied step,} \\[2pt]
\kappa\,\Delta s_t + b\,\mathbf{1}[\text{new checkpoint}], & \text{else,}
\end{cases}
\end{equation*}
with a terminal bonus $r_{goal}=+2.0$ on success and a terminal penalty
$r_{fail}=-0.10$ for falling in a hole, leaving the corridor, an
anti-cheat trip, or a timeout. The zero case covers a marble that is
occluded, off the path grid, or whose apparent jump was denied by the
anti-cheat check; such steps forfeit progress without directly penalizing
the policy. Otherwise, $\Delta s_t$ is the credited advance along the
discretized reference path, $\kappa$ converts it to reward, and a bonus
$b=0.02$ is paid once, the first time the marble reaches each new
checkpoint. A separate stuck check ends the episode as a failure if the
marble remains within a small radius for too long without progress.

\begin{figure*}[!t]
  \centering
  \includegraphics[width=0.90\textwidth]{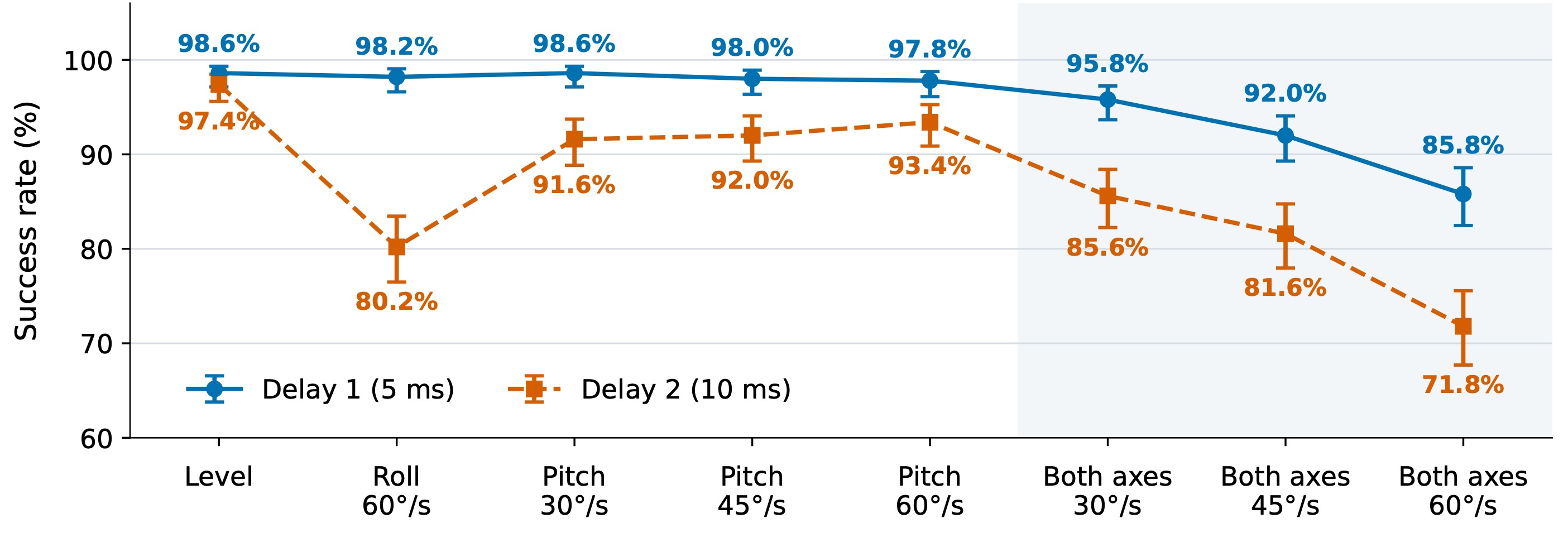}\vspace{-0.3cm}
  \caption{Simulation results: success rates of the trained policy under $5$ms and $10$ms action delays across a range of board-disturbance conditions. Error bars indicate $95\%$ Wilson confidence intervals.}
  \label{fig:pendulum-sim-success}\vspace{-0.3cm}
\end{figure*}
\subsection{Pendulum Control}\vspace{-0.1cm}
\label{sec:pendulum-simulation}

Reward is guarded against detector noise: a bad frame can make the marble
appear to jump implausibly far, which, credited at face value, would let
the policy farm reward from noise rather than genuine progress. A step's
progress is credited only if consistent with how far the marble could
plausibly have rolled in that time; larger jumps are denied credit until
the new position persists across several consecutive frames, so genuine
fast motion is still scored while a single bad detection is not.

\section{Simulation and Experimental Results}
\label{sec:simulation-experiments-results}
In this section, we present results from both simulation and real-hardware experiments on TAG. The pendulum and maze policies are trained separately, using fundamentally different data sources. The pendulum policy is trained entirely in simulation, with no adaptation or fine-tuning on real hardware. The maze policy, in contrast, is learned directly from real-hardware data: this data is used to train the DreamerV3 world model, and the corresponding actor-critic policy is then optimized using trajectories imagined within that learned latent space, without collecting any further real-world data during this optimization step.


The nominal simulated plant approximates
the physical pendulum using fixed parameters identified from hardware. 
Board motion is varied independently, over a broader disturbance envelope
than observed in the reported hardware experiment, which is our designed safety margin. As a result, the policy stays more robust and less surprised when
transferring to the physical platform which is well covered under simulated conditions.\vspace{0.1cm}

\noindent\textbf{Simulations Results.} We will evaluate the learned policy using simulation under diverse board disturbances and action delays for two cases: $5$ and $10$ ms delays. Action delay denotes the time between issuing a motor command and the onset of actuator response. Upright condition is defined as within $\pm10^\circ$ relative to true gravity upright reference.

For all simulations, we evaluate the trained policy over $500$ episodes per disturbance and delay settings. Success requires completing the episode and maintaining balance near its end, while satisfying the angular-velocity and cable-safety criteria. We report success rates with pointwise $95
\%$ Wilson confidence intervals \cite{newcombe1998two}, which give more reliable coverage than the standard normal-approximation interval, particularly for small sample sizes or success rates near $0\%$ or 
$100\%$.

Our simulation results are presented in Fig.~\ref{fig:pendulum-sim-success}. At $5$ ms delay, success under single-axis motion remains close to the stationary-board result of 
$98.6\%$, reaching  $98.2\%$ for roll and 
$97.8\%–98.6\%$ for pitch. Under simultaneous roll–pitch motion, success decreases from  $95.8\%$ to $92.0\%$ and $85.8\%$ as the speed limit increases.

At $10$ ms delay, the final policy retains $71.8\%–97.4\%$ success across the same conditions. Success on the stationary board is $97.4\%$. Under simultaneous roll–pitch motion, it decreases from $85.6\%$ to $81.6\%$ and $71.8\%$ as the speed limit increases. As the delay gets larger, the success rate reduces, but the controller continues to complete swing-up and stabilization throughout the evaluated envelope. These results demonstrate tolerance to delays beyond the measured hardware value. Across both delay settings, degradation is consistently more pronounced under simultaneous two-axis motion than under single-axis disturbance, indicating that the combined effect of longer delay and higher-dimensional disturbance compounds rather than acting independently. This suggests that fast, simultaneous roll–pitch motion under extended delay represents the most demanding regime within the evaluated envelope, and the one most relevant to sustained robustness during real coupled operation.

\begin{figure}[!t]
  \centering
  \includegraphics[width=\columnwidth]{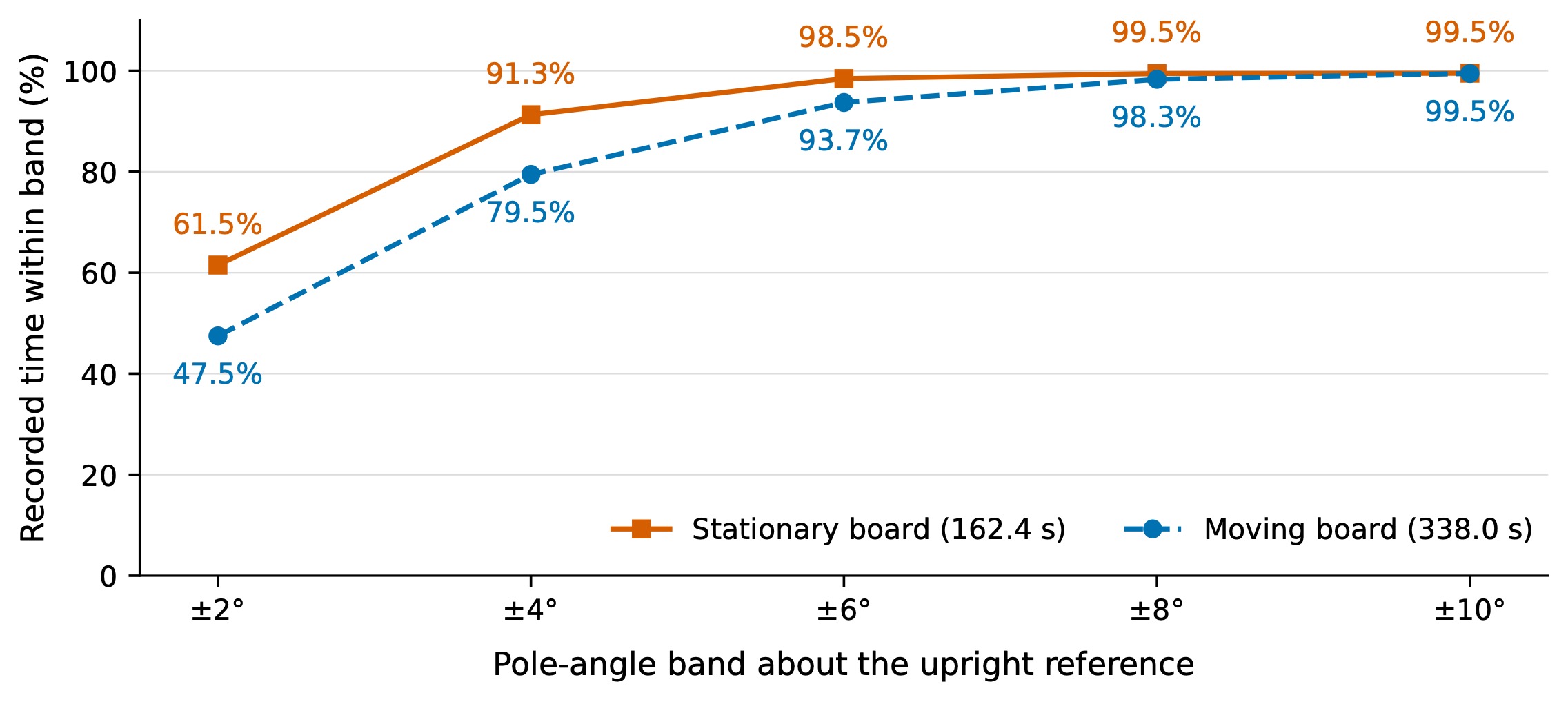}\vspace{-0.3cm}
  \caption{Real experimental results: success rates of the trained policy in real hardware across a range of board movements.}\vspace{-0.7cm}
  \label{fig:pendulum-hardware-bands}
\end{figure}
\noindent\textbf{Experimental Results on Hardware.} We implement the trained policy from simulation on real hardware in two settings:
$162.4$\,s with the board stationary, and $338.0$\,s with the board moving.
Pole angle is measured by the joint encoder relative to its board-fixed
upright reference. These experiments examine whether the disturbance
tolerance observed in simulation translates to the physical hardware. Our experimental results are shown in Fig. \ref{fig:pendulum-hardware-bands}. The stationary-board recording corresponds most closely to the level-board
simulation condition. In the moving-board recording, roll ranges from
$-4.48^\circ$ to $7.70^\circ$ and pitch from $-6.53^\circ$ to $5.67^\circ$,
both within the simulated $\pm10^\circ$ envelope, with peak absolute
gyroscope rates of $26.9^\circ$/s (roll) and $33.0^\circ$/s (pitch).
Notably, this measured disturbance is only about half the severity of the
hardest condition the policy was trained and evaluated against in
simulation, up to $60^\circ$/s. This gap is a deliberate consequence of our
training design, not a shortfall: training the policy against disturbance
more severe than what it typically encounters during real maze operation
allows the same policy to handle the easier, real-world case without
requiring any additional fine-tuning on hardware. Exposing the policy to
randomized uncertainty in friction, delay, and other parameters during
training further broadens the range of conditions it can tolerate, which we
expect contributes to this transfer. This robustness is not free, however;
it depends on first identifying the real system's parameters accurately
enough to center the randomization ranges correctly, since sufficiently
narrow, well-centered ranges are what let randomization improve robustness
without simply making the training distribution too wide to learn from
efficiently.

At the measured hardware delay of approximately $5$\,ms, the final policy
achieves $99.5\%$ occupancy within $\pm10^\circ$, compared with $95.8\%$
and $92.0\%$ for the simulated two-axis conditions at $30^\circ$/s and
$45^\circ$/s, respectively (Fig.~\ref{fig:pendulum-sim-success}). The
higher hardware success rate is consistent with the milder disturbance
actually observed on hardware: the real board does not move simultaneously
in roll and pitch at all times, and its operating range stays below the
peak envelope for most of the recording, so this comparison reflects a
genuinely easier operating condition on hardware rather than an unexpected
improvement over simulation.

Figure~\ref{fig:pendulum-hardware-bands} shows how closely the pendulum
tracks its upright reference in both recordings. On the stationary board,
occupancy reaches $61.5\%$ within $\pm2^\circ$, $91.3\%$ within $\pm4^\circ$,
and $98.5\%$ within $\pm6^\circ$. Under board motion, these values decrease
to $47.5\%$, $79.5\%$, and $93.7\%$, respectively. The gap narrows as
tolerance widens: occupancy within $\pm8^\circ$ is $99.5\%$ for the
stationary recording and $98.3\%$ for the moving recording, and both reach
approximately $99.5\%$ within $\pm10^\circ$.

Board motion measurably reduces fine regulation precision in these
recordings, while the pendulum remains within the wider upright band for
nearly all of the active operation. This behavior is consistent with the
disturbance tolerance observed in simulation and supports the practical
transfer of the trained controller to the physical platform. Taken
together, these hardware results suggest that the curriculum, system
identification, and domain randomization used during training generalize
beyond the specific disturbance levels directly tested in simulation, and
that evaluating a policy under conditions harder than expected deployment
is a useful, low-cost way to build in a margin of robustness for
real-world operation without requiring hardware-in-the-loop fine-tuning.\vspace{-0.1cm}

\subsection{Ball-in-Maze Control}\vspace{-0.1cm}
\label{sec:coupling-experiment}
We use a single continuous training run on the physical system, structured incrementally rather than training against the fully coupled system from the start. The maze policy first trains for 
$9,741$ episodes with the pendulum inactive: the two servos tilt the board to move the camera-tracked marble along a fixed route, with no pendulum-induced motion present. At episode
$9,741$, the pendulum is switched on and training continues with the same policy, without resetting or retraining it from scratch. Training directly under pendulum-induced disturbance from a randomly initialized policy would confront it with two difficult problems at once: learning the maze route itself, and tolerating an additional, physically-coupled disturbance, with little useful signal for either until the marble can be steered reliably. Learning the maze undisturbed first lets the policy establish a competent world model and control strategy before the pendulum is introduced. Introducing the pendulum mid-training, rather than in a separate policy or a fresh run, directly tests the central question of this experiment: how an already-competent policy responds to a newly introduced physical disturbance from an independently operating subsystem.



\begin{figure}[t]
  \centering
  \includegraphics[width=\columnwidth]{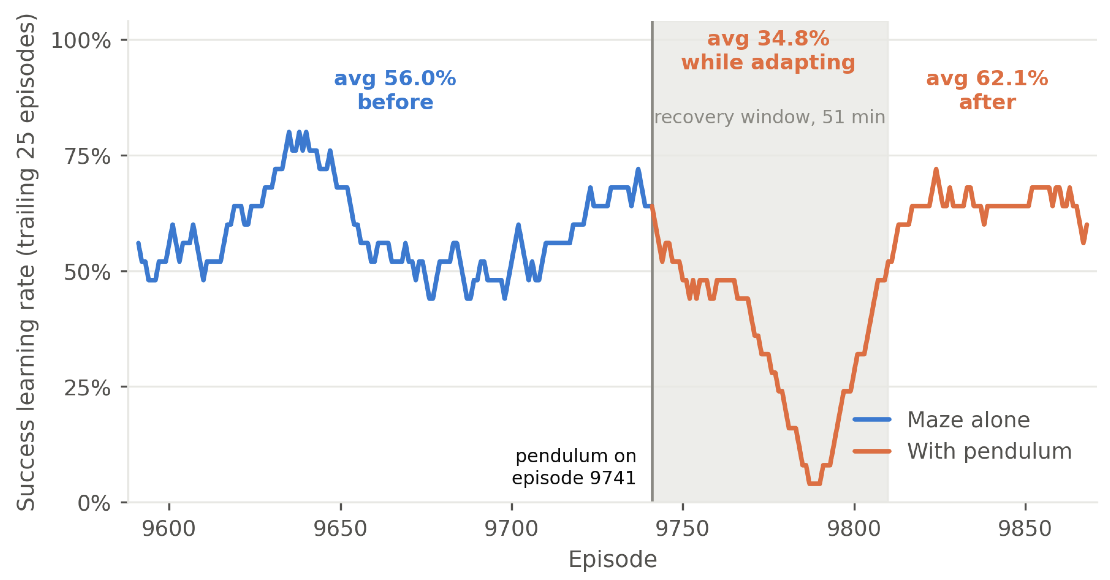}\vspace{-0.3cm}
  \caption{Incremental training of the maze policy before and after pendulum activation.  Blue and orange denote training with the inactive and active pendulum, respectively, and the shaded region marks the 69-episode ($\sim$51 minutes) adaptation window under pendulum activation at episode 
9,741.}\vspace{-0.6cm}
  \label{fig:coupling-success}
\end{figure}

\noindent\textbf{Experimental Results.} Figure~\ref{fig:coupling-success} reports the success rate observed during
learning, i.e., how often the marble reached the goal in training episodes.
The blue curve shows performance while the pendulum is inactive; the
orange curve shows performance once the pendulum is switched on and
running under its own controller. The shaded region marks the adaptation
window, spanning the transition from the pendulum-inactive to
pendulum-active condition. Each point is a 25-episode average, not the
result of a single episode.

In the 300 episodes immediately preceding activation (episodes
$9{,}441$--$9{,}741$), training success is $56.0\%$. Over the following 69
episodes, success decreases to $34.8\%$, before recovering to $62.1\%$
over the subsequent 58 episodes, a level that exceeds the pre-activation
baseline. The training data therefore show an immediate performance cost
upon activation, followed by a return to, and slight improvement over, the
prior success level within 69 episodes, roughly 51 minutes of continued
training. Because the policy continues to update throughout this run,
these phase averages describe an ongoing adaptation process rather than
the performance of fixed, frozen policy checkpoints. This
experiment shows fast online adaptation to a newly introduced
physical disturbance.

\noindent\textbf{Main Observations.} First, initial learning required $9{,}741$ episodes, whereas adapting to
the activated pendulum required 69 episodes. Comparing these episode counts
gives $9{,}741/69\approx140$: adaptation was approximately
$140\times$ faster in episodes and took 51 minutes. This result is consistent
with the world model and policy retaining the maze task while updating their
representation of the added disturbance.

Second, at its lowest point, the trailing success curve reaches approximately
$4\%$, a single point on the 25-episode moving average. Therefore, it is not directly
comparable to the $34.8\%$ figure, which averages over the complete
69-episode window. This implies that the sharp, momentary dip to
$4\%$ indicates the pendulum's introduction is initially highly disruptive,
while the comparatively higher 69-episode average shows the policy begins
recovering well before the full adaptation window has elapsed.

Episode-level behavior shows the collapse is concentrated at the start of
episodes, not spread across the route: median episode duration during the
dip is only $0.7$\,s, compared with $23.7$\,s across the 50 episodes
immediately preceding activation, and 20 of 30 episodes in the dip window
end in under $5$\,s. This implies failures cluster immediately after
reset, when the marble's state is least predictable and the policy has
had the least opportunity to establish control. The automatic reset does
not place the marble at rest; it arrives with nonzero velocity and spin,
and pendulum motion compounds this unstable initial condition by altering
the marble's contact behavior and response to board tilt, in some cases
carrying it into a hole before the policy can re-establish control. This
is consistent with the pendulum's disturbance being most costly where the
maze policy has the least established control authority, rather than
degrading performance uniformly throughout the route.

\section{Future Works}\vspace{-0.1cm}
The pendulum, trained offline in simulation with domain randomization,
degrades gracefully under disturbance ($98.6\%\to85.8\%$), while the
maze policy, trained online directly on hardware, collapses sharply
upon pendulum activation ($56.0\%\to\sim4\%$) before recovering to
$62.1\%$ within 69 episodes ($\sim$51 minutes), exceeding its
pre-activation level. This contrast suggests real-time exposure to
coupling can act as an implicit source of robustness and adaptation. Our approach also reflects
two distinct strategies for costly real-world interaction: avoiding
hardware data entirely via sim-to-real transfer, versus learning
directly on hardware with a world model that makes real episodes more
sample-efficient. Both prove effective: the two policies, trained
separately with no access to each other's model, each remain effective
despite disturbance transmitted from the other. Whether these
observations reflect genuine adaptation and
robustness remains open for future work under more
diverse conditions.

The maze-pendulum coupling can be framed as a leader-follower Stackelberg game, with the board as leader and the pendulum as follower, since the board's actions directly shape the pendulum's disturbance while the reverse influence is comparatively indirect. This would let Stackelberg learning algorithms be applied and tested directly on physical hardware, and would let us compare whether game-theoretic training produces different joint behavior than the independently-trained policies studied here.

Finally, TAG's modular design supports multi-task learning across multiple boards, pendulum configurations, and environmental conditions. Training across this variation would test whether policies adapt quickly to new physical instances rather than overfitting to one setup, and would help establish TAG as a multi-instance benchmark where fast adaptation, not just single-platform performance, is the object of study.\vspace{-0.1cm}
\bibliographystyle{IEEEtran}
\bibliography{refs}







\end{document}